\documentclass{article}
\usepackage[preprint]{neurips_2025}

\usepackage[utf8]{inputenc} 
\usepackage[T1]{fontenc}    
\usepackage{hyperref}       
\usepackage{url}            
\usepackage{booktabs}       
\usepackage{amsfonts}       
\usepackage{nicefrac}       
\usepackage{microtype}      
\usepackage{xcolor}         
\usepackage{arydshln}         
\usepackage{comment}
\usepackage{graphicx}
\usepackage{amssymb}
\usepackage{svg}
\usepackage{amsmath}
\usepackage{tabularx}
\usepackage{siunitx}
\usepackage{subscript}
\usepackage{pifont}
\usepackage{tikz}
\usepackage{multirow}
\usepackage{float}
\usepackage{colortbl}
\usepackage{enumitem}
\usepackage{latexsym}
\usepackage{amssymb}
\usepackage{arydshln}
\usepackage{makecell}
\usepackage{wrapfig}

\usepackage{amsmath}
\usepackage[most]{tcolorbox}
\newtcolorbox{takeawaybox}[2][]{
  enhanced,
  colback=blue!5!white,
  colframe=black,
  coltitle=white,
  fonttitle=\bfseries,
  colbacktitle=black,
  boxed title style={
    rounded corners,
    boxrule=0pt,
    arc=1mm
  },
  title style={parindent=0pt},
  attach boxed title to top left={yshift=-2mm,xshift=2mm},
  title=#2,
  #1
}

\newcolumntype{C}[1]{>{\centering\arraybackslash}p{#1}}
\title{KDRL: Post-Training Reasoning LLMs via Unified Knowledge Distillation and Reinforcement Learning}

\author{
Hongling Xu$^{1}$\thanks{\quad This work was done during the internship at Huawei Noah’s Ark Lab.}~~, Qi Zhu$^{2}$\thanks{\quad Corresponding authors.}~~,
  Heyuan Deng$^{2}$, Jinpeng Li$^{2}$,
  Lu Hou$^{2}$,  
  \\
  \bf Yasheng Wang$^{2}$, Lifeng Shang$^{2}$, Ruifeng Xu$^{1}$\footnotemark[2]~~, Fei Mi$^{2}$
  \\ [1mm]
  $^{1}$ Harbin Institute of Technology (Shenzhen) \quad
  $^{2}$ Huawei Noah’s Ark Lab \\ [0.5mm]
  \texttt{xuhongling@stu.hit.edu.cn},~\texttt{zhuqi41@huawei.com},~\texttt{xuruifeng@hit.edu.cn}
}

\begin{document}

\maketitle

\begin{abstract}
Recent advances in large language model (LLM) post-training have leveraged two distinct paradigms to enhance reasoning capabilities: reinforcement learning (RL) and knowledge distillation (KD). 
While RL enables the emergence of complex reasoning behaviors, it often suffers from low sample efficiency when the initial policy struggles to explore high-reward trajectories. 
Conversely, KD improves learning efficiency via mimicking the teacher model but tends to generalize poorly to out-of-domain scenarios.
In this work, we present \textbf{KDRL}, a \textit{unified post-training framework} that jointly optimizes a reasoning model through teacher supervision~(KD) and self-exploration~(RL).
Specifically, KDRL leverages policy gradient optimization to simultaneously minimize the reverse Kullback–Leibler divergence (RKL) between the student and teacher distributions while maximizing the expected rule-based rewards.
We first formulate a unified objective that integrates GRPO and KD, and systematically explore how different KL approximations, KL coefficients, and reward-guided KD strategies affect the overall post-training dynamics and performance.
Empirical results on multiple reasoning benchmarks demonstrate that KDRL outperforms GRPO and various KD baselines while achieving a favorable balance between performance and reasoning token efficiency.
These findings indicate that integrating KD and RL serves as an effective and efficient strategy to train reasoning LLMs.
\end{abstract}

\section{Introduction}

\begin{figure*}[t]
    \centering
    \includegraphics[width=0.85\textwidth]{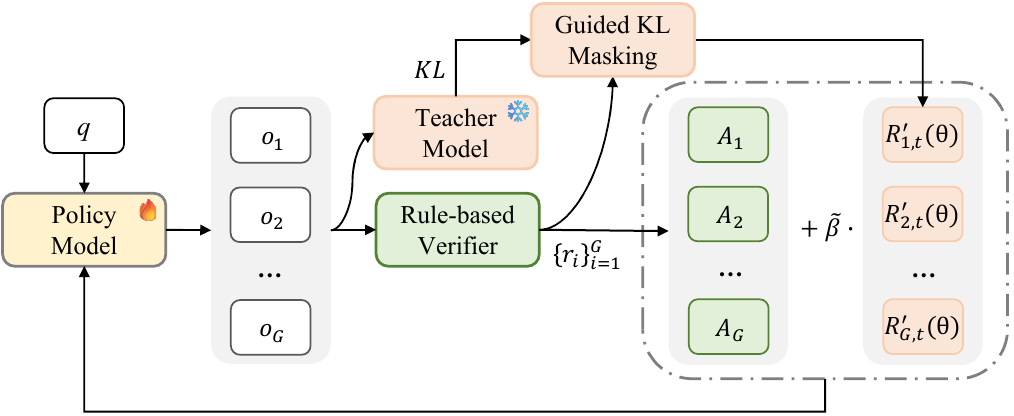}
    \caption{Overview of the proposed \textbf{KDRL}, a unified framework that combines knowledge distillation (orange) and reinforcement learning (green) for post-training reasoning LLMs. The two components are integrated via policy gradient optimization: a rule-based verifier scores on-policy rollouts to provide reward signals, while a teacher model offers token-level supervision through KL divergence. Additionally, we explore KL approximation, KL scheduling, and reward-guided distillation to enhance the effectiveness and efficiency of KDRL.}
    \vspace{-2 mm}
    \label{fig:overview}
\end{figure*}

Post-training serves as an important technique for enhancing the reasoning capabilities of large language models (LLMs), and \emph{reinforcement learning} (RL) and \emph{knowledge distillation} (KD) are two commonly used techniques. Both OpenAI o-series models~\cite{jaech2024openai} as well as DeepSeek-R1~\cite{guo2025deepseek} reveal that scaling RL during post-training can effectively incentivize and boost reasoning capability in LLMs, while DeepSeek-R1's experiments also show that distilling from a powerful teacher model yields better results than RL for less capable models.
Distilling knowledge from a more powerful teacher model using supervised fine-tuning (SFT)~\cite{ouyang2022instruction} or rejection sampling fine-tuning~\cite{yuan2023rft} (RFT) has been widely used for LLM post-training.
KD is effective and efficient in the sense that it can learn from teacher supervision, but it is constrained by the capability of the teacher and often falls short in out-of-domain generalization~\cite{chu2025sft}.
In contrast, RL can amplify useful reasoning patterns within the model itself through self-exploration and reward guidance, often leading to better generalization. 
However, it is constrained by the base model’s inherent capacity~\cite{gandhi2025cognitive,yue2025does} and requires substantially more computation to search and optimize. 
Therefore, existing strategies~\cite{guo2025deepseek,ouyang2022instruction,grattafiori2024llama3,yang2024qwen25} often apply KD first to improve the model's capability and then further refine the model via RL in two or even more separate stages.
However, these separate paradigms may hinder the scaling and efficiency of post-training as teacher supervision and self-exploration are decoupled.
To take advantage of both teacher supervision and self-exploration for better post-training efficiency and effectiveness, we propose \textbf{KDRL}, a unified framework that jointly optimizes KD and RL, as illustrated in Figure~\ref{fig:overview}.
For KD, rather than the traditional SFT that minimizes a standard cross-entropy loss on teacher outputs, we minimize the reverse Kullback–Leibler divergence (RKL)~\cite{gu2024minillm,agarwal2024onpolicy} between the policy model and the teacher model using on-policy rollouts.
We refer to this approach as KD-RKL, which allows seamless integration of RL algorithms (e.g., GRPO) with the distillation objective.

In our study, we theoretically and empirically study the effects of 
(i) different KL approximation methods, 
(ii) weights and scheduling techniques for balancing the contribution of KD and RL, and
(iii) dynamic masking strategies to select ``hard'' data samples to be optimized through teacher supervision in our KDRL.
We highlight that the $k2$ KL estimator, a balanced combination of KD and RL objectives, and response-level KL masking are each important factors contributing to the overall performance or token efficiency of KDRL. 
Furthermore, we conduct extensive experiments comparing KDRL with GRPO and various KD baselines.
We observe that KDRL largely outperforms standard SFT approaches and GRPO by 5.7\% and 2.6\%, respectively, and also outperforms the on-policy distillation approach KD-RKL, which is similar to a contemporaneous implementation in Qwen3~\cite{yang2025qwen3technicalreport}, by 1.1\%.
Apart from effectiveness, we also demonstrate that KDRL achieves a favorable trade-off between performance and efficiency in both training cost and reasoning token usage, making it a promising approach for scaling LLM post-training.

\section{Preliminary}

\subsection{Reinforcement Learning}
In RL for LLMs, policy gradient methods are widely used, which aim to optimize the policy directly using gradient ascent.
Particularly, the objective of the classic REINFORCE method~\cite{williams1992simple} is
\begin{align}
\mathcal{J}_{\text{REINFORCE}}(\theta)=\mathbb{E}_{q\sim Q, o\sim \pi_{\theta}(\cdot|q)}[r(o,q)], \label{eq:reinforce}
\end{align}
where $q,o$ are questions and responses sampled from the question dataset $Q$ and the policy model $\pi_\theta$, respectively. $r(o,q)$ is the reward function.
To improve training stability and efficiency, two key techniques are introduced: replacing the reward function with the advantage function to reduce gradient variance, and constraining the magnitude of policy updates.
Representative algorithms include PPO~\cite{schulman2017proximal} and GRPO~\cite{shao2024deepseekmath}.
We adopt GRPO with outcome reward as the RL algorithm in our study due to its computational efficiency. Additionally, we incorporate two modifications to enhance performance, namely, \textit{removing KL constraint} and \textit{leveraging token-level loss}, following~\cite{yu2025dapo,skywork-or1-2025,liu2025understanding}.
GRPO samples a group of responses for each question from the old policy $\pi_{\theta_{old}}$. We adopt on-policy optimization by setting $\pi_{\theta_{\text{old}}} = \pi_\theta$, and then optimize the following objective:
\begin{align}
    \mathcal{J}_{\text{GRPO}}(\theta) &= 
    \mathbb{E}_{q \sim Q, \{o_i\}^G_{i=1} \sim \pi_{\theta_{\text{old}}}(\cdot | q)}  
    \bigg[ \frac{1}{\sum_{i=1}^G|o_i|} \sum_{i=1}^{G} \sum_{t=1}^{|o_i|} 
    \rho_{i,t}(\theta)\hat{A}_{i} \bigg], \label{eq:GRPO_NEW}
\end{align}
where $\hat{A}_i=\frac{r_i-\text{mean}(\{r_1,r_2,..,r_G\})}{\text{std}(\{r_1,r_2,..,r_G\})}$ is the advantage function, and $r_i$ is the outcome reward of trajectory $o_i$. $\rho_{i,t}{(\theta)}=\frac{\pi_{\theta}(o_{i,t}|q,o_{i,< t})}{\pi_{\theta_{old}}(o_{i,t}|q,o_{i,< t})}$ is the weight of importance sampling.

\subsection{Knowledge Distillation}
Knowledge distillation aims to train the student policy $\pi_\theta$ to mimic the behavior of a more powerful teacher $\pi_T$~\cite{hinton2015distilling,gou2021knowledge,xu2024survey}. 
A simple yet effective method is to maximize the log-likelihood on data generated by the teacher, also known as supervised fine-tuning (SFT)~\cite{guo2025deepseek,hsieh-etal-2023-distilling,abdin2024phi}. It is equivalent to minimizing the forward KL divergence between $\pi_T$ and $\pi_\theta$:
\begin{align}
\arg \min_\theta \mathbb{D}_{\text{KL}}(\pi_T||\pi_\theta)=\arg \max_\theta \mathbb{E}_{q\sim Q, o\sim \pi_T(\cdot|q)}[\log\pi_\theta(o|q)]=
    \arg \max_\theta \mathcal{J}_{\text{SFT}}(\theta).
\end{align}
However, training on such off-policy data induces exposure bias~\cite{bengio2015scheduled}: a mismatch between training on teacher-generated prefixes and making predictions on self-generated prefixes, which is especially severe for reasoning LLMs due to their long responses.
A way to alleviate this issue is training on self-generated prefixes, which is equivalent to minimizing the reverse KL divergence (RKL)~\cite{gu2024minillm}:
\begin{align}
    \arg \min_\theta \mathbb{D}_{\text{KL}}(\pi_\theta||\pi_T) &= \arg \max_\theta \mathbb{E}_{q\sim Q,\, o\sim \pi_\theta(\cdot|q)}\bigg[\log\frac{\pi_T(o|q)}{\pi_\theta(o|q)}\bigg].
\end{align}
Compared with SFT, KD with RKL performs on-policy sampling, which requires more computational cost but results in lower exposure bias and better performance~\cite{gu2024minillm,agarwal2024onpolicy,team2024gemma, ko2024distillm}.

Interestingly, we find that minimizing RKL can be regarded as REINFORCE in Eq.~\eqref{eq:reinforce} with $\log\frac{\pi_T(o|q)}{\pi_\theta(o|q)}$ as the reward function: encouraging trajectories that the teacher prefers more and discouraging those it deems unlikely.
Therefore, adopting the same techniques as GRPO except the advantage function, we can optimize the following objective ($\pi_{\theta_{old}}=\pi_\theta$ as above):
\begin{align}
\mathcal{J}_{\text{RKL}}(\theta) &= 
    \mathbb{E}_{q \sim Q, \{o_i\}^G_{i=1} \sim \pi_{\theta_{\text{old}}}(\cdot | q)}  
    \bigg[ \frac{1}{\sum_{i=1}^G|o_i|} \sum_{i=1}^{G} \sum_{t=1}^{|o_i|} 
    \rho_{i,t}(\theta)R_{i,t}(\theta) \bigg],
\label{eq:RKL}
\end{align}
where the log-likelihood ratio $R_{i,t}(\theta) = \log \frac{\pi_T(o_{i,t} \mid q, o_{i,<t})}{\pi_\theta(o_{i,t} \mid q, o_{i,<t})}$ acts as the token-level reward signal. When $R_{i,t}(\theta)>0$, it encourages the student to increase the probability of the current token, and conversely decrease it.

\subsection{KL Approximation}
The KL divergence $\mathcal{J}_{\text{RKL}}(\theta)$ is computed by Monte Carlo sampling.
Due to the high variance of the direct KL estimator $R_{i,t}(\theta)$, $k2$ and $k3$ KL approximations are proposed~\cite{schulman_kl_approx}. In our context, the $k2$ and $k3$ KL can be calculated as follows:
\begin{align}
\mathbb{D}_{\mathrm{KL}}^{k2}(\pi_\theta \| \pi_T) &=\mathbb{E}_{q \sim Q, \{o_i\}^G_{i=1} \sim \pi_{\theta_{\text{old}}}(\cdot | q)}  
    \bigg[ \frac{1}{\sum_{i=1}^G|o_i|} \sum_{i=1}^{G} \sum_{t=1}^{|o_i|} 
    \frac{1}{2} R_{i,t}^2(\theta)\bigg].\\
\mathbb{D}_{\mathrm{KL}}^{k3}(\pi_\theta \| \pi_T) &=\mathbb{E}_{q \sim Q, \{o_i\}^G_{i=1} \sim \pi_{\theta_{\text{old}}}(\cdot | q)}  
    \bigg[ \frac{1}{\sum_{i=1}^G|o_i|} \sum_{i=1}^{G} \sum_{t=1}^{|o_i|} 
    (e^{R_{i,t}(\theta)} - R_{i,t}(\theta) - 1 )\bigg].
    \label{eq:approx-kl}
\end{align}
Although $\mathbb{D}_{\mathrm{KL}}^{k3}(\pi_\theta \,\|\, \pi_T)$ is an unbiased estimator of $-\mathcal{J}_{\mathrm{RKL}}(\theta)$, its gradient $\nabla_\theta \mathbb{D}_{\mathrm{KL}}^{k3}(\pi_\theta \,\|\, \pi_T)$ is biased with respect to $-\nabla_\theta \mathcal{J}_{\mathrm{RKL}}(\theta)$. In contrast, while $\mathbb{D}_{\mathrm{KL}}^{k2}(\pi_\theta \,\|\, \pi_T)$ itself is a biased estimator of $-\mathcal{J}_{\mathrm{RKL}}(\theta)$, its gradient $\nabla_\theta \mathbb{D}_{\mathrm{KL}}^{k2}(\pi_\theta \,\|\, \pi_T)$ provides an unbiased estimate of $-\nabla_\theta \mathcal{J}_{\mathrm{RKL}}(\theta)$.
See the full derivation in Appendix \ref{appendix:math}.

Another way to reduce the estimation variance of $\mathcal{J}_{\text{RKL}}(\theta)$ is to compute the discrepancy between the student and teacher \textit{over the full vocabulary}~\cite{gu2024minillm,sanh2019distilbert} instead of on sampled token only.
However, storing full distributions imposes prohibitive memory overhead. 
A practical approach is the Top-$K$ approximation~\cite{peng2024pretrainingdistillationlargelanguage,li-etal-2025-bild}.
We retain only the top-$K$ logits from the teacher prediction and re-normalize them to form a teacher distribution $\pi_T'$ for full vocabulary KL computation (See Appendix \ref{appendix:math}).

\section{KDRL: Unifying KD and RL for Post-training Reasoning LLMs}

Given the advantages of RKL in KD and its alignment with the RL objective, we use RKL as the KD objective to combine KD and RL.
We propose KDRL, a framework for training the model with teacher supervision and environment feedback simultaneously.
In the following sections, we thoroughly investigate 
(i) methods for integrating the KD-RKL and RL objectives, 
(ii) various KD-RKL estimators, 
(iii) the influence of the KD-RKL coefficient, and 
(iv) sample-masking strategies for KD-RKL computation.
The experimental setup is detailed in Section \ref{sec:setting}.

\subsection{Combination of GRPO and KD-RKL}
\label{combination}

In this section, we investigate two ways of integrating KD and RL, using \textit{reward shaping} and \textit{joint loss}, respectively.

\textbf{Reward Shaping.}  
This approach augments the original outcome reward $r_i$ by incorporating token-level KL rewards:
$
r_i' = r_i + \beta\sum_{t=1}^{|o_i|}  R_{i,t}(\theta).
$
\footnote{We use $\beta$ as 2e-3 unless otherwise specified. The effect of the KL coefficient is explored in Section \ref{sec:KL_coefficient}.} This modified reward is then used to calculate the estimated advantage $\hat{A}_i$ and update the policy as GRPO.

\textbf{Joint Loss.} 
Instead of combining $\mathcal{J}_{\text{GRPO}}(\theta)$ and $\mathcal{J}_{\text{RKL}}(\theta)$ (Eq. \ref{eq:RKL}) directly, this strategy augments the GRPO objective with an auxiliary $k2$ KL loss term $\mathbb{D}_{\mathrm{KL}}^{k2}(\pi_{\theta}\|\pi_{T})$:
\begin{align}
    \mathcal{J}_{\text{KDRL}}(\theta)
    &= \mathcal{J}_{\text{GRPO}}(\theta) 
    - \beta \mathbb{D}_{\mathrm{KL}}^{k2}(\pi_{\theta}\|\pi_{T}).
    \label{eq:GRPO-KD_grad}
\end{align}
Its gradient $\nabla_\theta\mathcal{J}_{\text{KDRL}}(\theta)$ is an unbiased estimate of $\nabla_\theta \mathcal{J}_{\text{GRPO}}(\theta) +  \beta\cdot\nabla_\theta \mathcal{J}_{\text{RKL}}(\theta)$.

\begin{figure*}[t]
    \centering
    \includegraphics[width=0.97\textwidth]{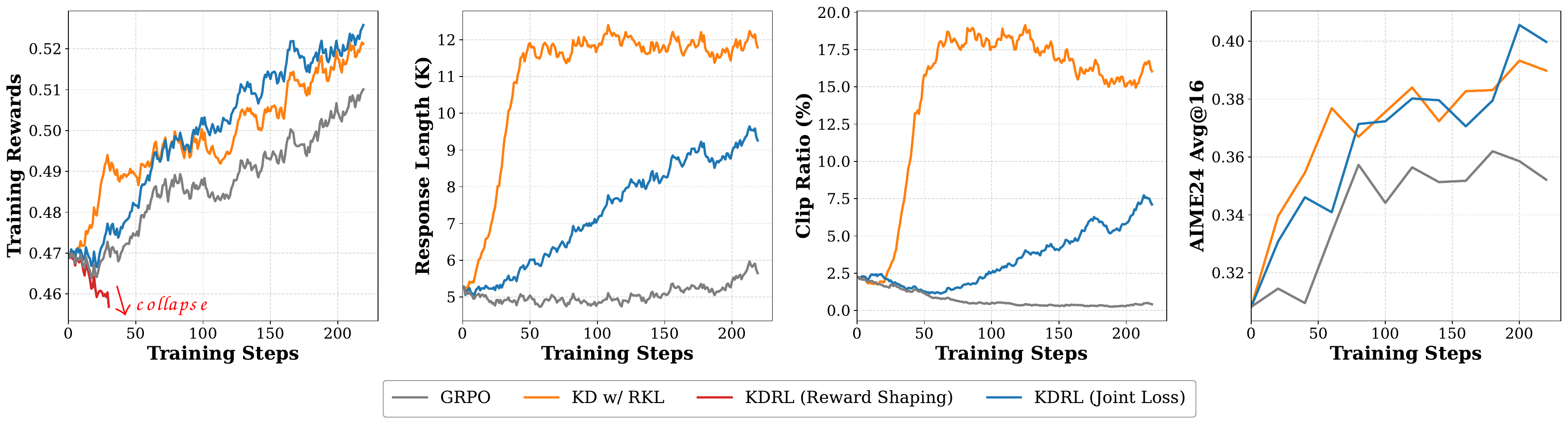}
    \vspace{-1mm}
    \caption{Training dynamics and performance of GRPO, KD-RKL, and KDRL variants with different KD integration strategies. From left to right: (a) training rewards, (b) average response length, (c) response clip ratio, and (d) AIME24 accuracy. All curves are smoothed using EMA. }
    \label{fig:combination}
\end{figure*}

\textbf{Results.}  
As shown in Figure~\ref{fig:combination}, among the two integration strategies, \textit{reward shaping} leads to collapsed training dynamics in early steps. In contrast, \textit{joint loss} yields stable improvements: the resulting KDRL method outperforms both GRPO and KD-RKL in terms of reward and validation accuracy, demonstrating the benefit of combining distillation and reinforcement learning. 
Additionally, we observe distinct trends in length variation. GRPO leads to gradual length growth, while KD-RKL causes rapid inflation and high truncation rates (approaching 15\%). KDRL with Joint Loss, however, has a more stable growth, producing moderately long responses that balance generation efficiency and performance.

\begin{takeawaybox}{Takeaway 3.1 - Combination of GRPO and KD-RKL}
While the combination of GRPO and KD-RKL through reward shaping collapses, their integration via a joint loss outperforms both GRPO and KD-RKL.
\end{takeawaybox}

\subsection{Approximations of KL Divergence}
\begin{figure*}[t]
    \centering
    \includegraphics[width=0.97\textwidth]{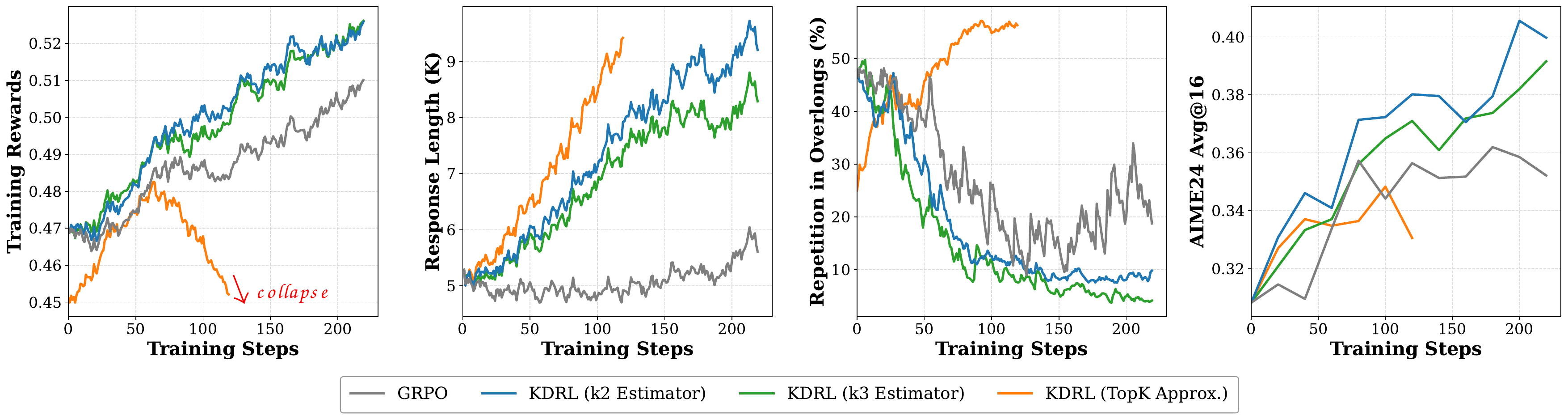}
    \vspace{-1mm}
    \caption{Comparison of different KL approximations. From left to right: (a) training rewards, (b) average response length, (c) repetition rate in overlong responses, and (d) AIME24 accuracy.}
    \vspace{-2mm}
    \label{fig:approximation}
\end{figure*}
Given the observed benefits of KDRL through loss combination, we further examine alternative formulations of RKL loss other than $\mathbb{D}_{\mathrm{KL}}^{k2}$. 
We explore three approximations and compare their effectiveness in KDRL: $k2$, $k3$, and Top-$K$.

\textbf{Results.}
Figure~\ref{fig:approximation} compares the KDRL performance across different KL approximation strategies. 
Both the $k2$ and $k3$ estimators yield similar training rewards, but the unbiased $k2$ estimator consistently achieves better performance on downstream tasks, with a 1.4\% improvement in AIME24 accuracy over $k3$.
In contrast, the Top-$K$ approximation leads to unstable training, as the reward drops early and no accuracy gain is observed. 
This instability likely stems from excessive KL penalties incurred by tokens outside the top-$K$ set, which destabilize the model's output. 
As shown in the third plot, a sharp increase in the repetition ratio within overlong responses is observed. 
Furthermore, we observe that unbiased estimators $k2$ tend to accelerate the growth of response length, likely due to faster convergence towards the teacher distribution.

\begin{takeawaybox}{Takeaway 3.2 - KL Approximation Methods}
In KDRL, the unbiased gradient estimator $k2$ outperforms the commonly used $k3$, and the Top-$K$ approximation leads to unstable optimization.
\end{takeawaybox}

\begin{figure*}[t]
    \centering
    \includegraphics[width=0.97\textwidth]{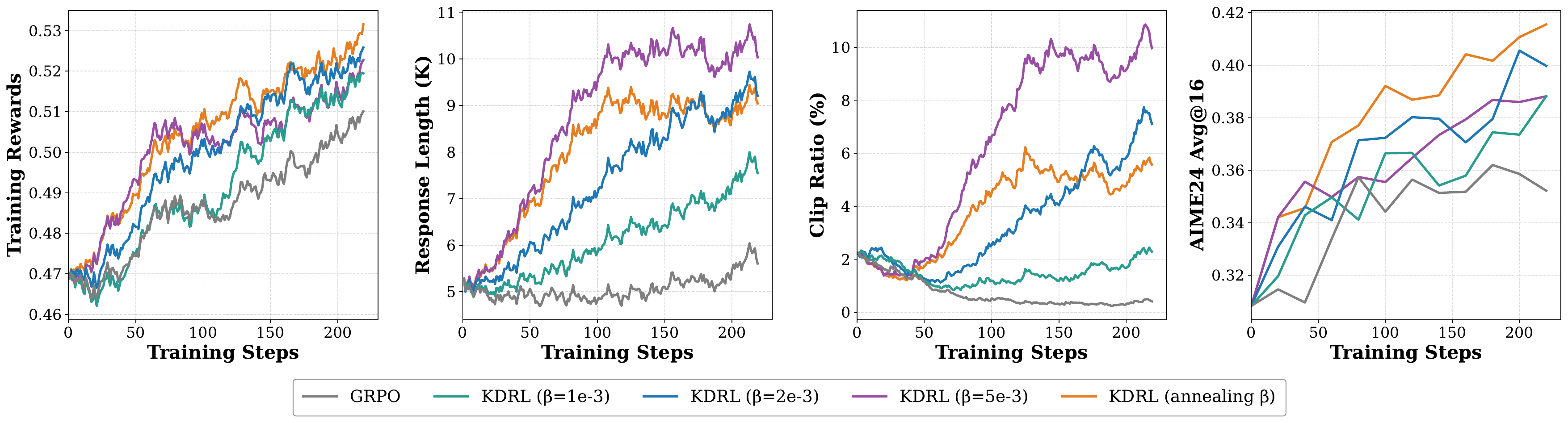}
    \caption{Comparison of different KL coefficient settings. From left to right:  
(a) training rewards, (b) average response length, (c) response clip ratio, and (d) AIME24 accuracy. }
    \vspace{-2 mm}
    \label{fig:coefficient}
\end{figure*}

\subsection{Balance of RL and KD}
\label{sec:KL_coefficient}

In the KDRL framework, the KL coefficient $\beta$ governs the trade-off between the RL reward optimization (via GRPO) and teacher imitation (via KD-RKL).  
As discussed in Section~\ref{combination}, relying solely on KD tends to cause early length inflation and high truncation rates, while naïve GRPO results in slower growth in both reasoning length and task performance. These observations highlight the importance of tuning $\beta$ to balance the two learning signals.
Motivated by this, we examine the impact of different KL coefficient settings, including both constant and dynamic scheduling.

\textbf{Constant KL Coefficient.}  
In addition to the default value of $\beta=$2e-3, we evaluate two additional settings, 1e-3 and 5e-3, to assess how the strength of distillation influences learning dynamics. A smaller $\beta$ allows the policy to rely more on reward signals, while a larger $\beta$ enforces stronger teacher supervision.

\textbf{Annealing KL Coefficient.}  
While strong KD signals can offer guidance in the early stages, maintaining a high $\beta$ all the time may hinder the model's continual improvement.
Therefore, we adopt a linearly decaying KL schedule:
$\beta = \max(\beta_{\text{init}} - \delta \cdot \text{step},\ \beta_{\text{min}})$.
In our experiments, we set $\beta_{\text{init}}$ to 5e-3, $\delta$ to 5e-5, and $\beta_{\text{min}}$ to 1e-3. This schedule facilitates stronger teacher supervision early on, followed by a gradual shift toward reward-driven optimization.

\textbf{Results.}
Figure~\ref{fig:coefficient} illustrates the impact of different KL coefficient settings. We observe that increasing $\beta$ strengthens teacher supervision, as indicated by longer generated responses. However, a relatively large $\beta$ (e.g., 5e-3 in our setting) results in rapid length inflation and lower reward acquisition, likely due to increased truncation. 
In contrast, a smaller $\beta$ such as 1e-3 yields more stable length growth but underutilizes teacher supervision. Notably, the linearly annealed $\beta$ schedule achieves the best overall performance, which matches the length of the 2e-3 setting while maintaining a lower clip ratio, suggesting more effective reward exploitation. These results indicate that gradually shifting from strong imitation to reward-driven optimization leads to more effective training.

\begin{takeawaybox}{Takeaway 3.3 - Balance RL and KD}
Both moderate KL coefficient and annealing schedule help balance KD and RL. Annealing further improves performance by a smooth transition from imitation to reward optimization.
\end{takeawaybox}

\subsection{Reward-Guided KD}
\label{sec:masking}
\begin{figure*}[t]
    \centering
    \includegraphics[width=0.97\textwidth]{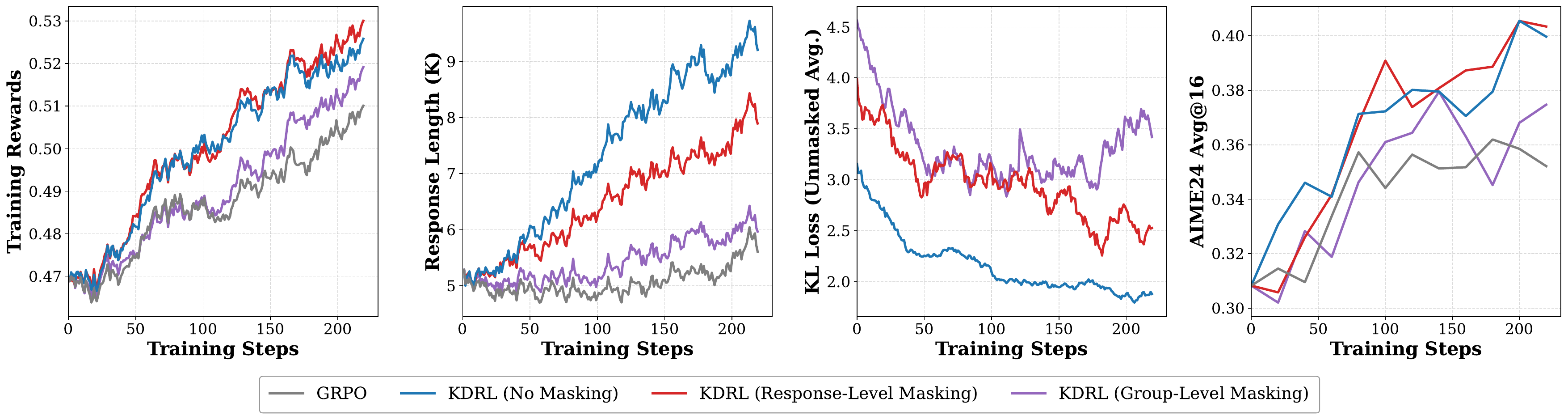}
    \caption{Comparison of baselines and  KDRL with response-level and group-level KL masking. From left to right: (a) training rewards, (b) average response length, (c) KL loss (computed on unmasked tokens), and (d) AIME24 accuracy.}
    \vspace{-2mm}
    \label{fig:masking}
\end{figure*}

In outcome-supervised GRPO, the advantage is computed at the response level and then distributed across tokens. However, when combined with KD, this setup can introduce conflicting gradient signals: even for correct responses, the RKL objective penalizes tokens that diverge from the teacher, potentially interfering with reward-aligned learning and destabilizing training.
To address this issue, we propose \textit{Reward-Guided KL Masking}, which conditions the KD-RKL loss on reward feedback. The core idea is that selective imitation suffices: KD could be suppressed for responses or groups that already receive positive feedback, allowing distillation to focus on regions where the policy underperforms. We explore two levels of masking granularity:

\textbf{Response-Level Masking.}  
We apply masking at the individual response level: if a response receives a positive reward, its KD loss is omitted. Specifically, we replace the KD signal \( R_{i,t}(\theta) \) by \( R'_{i,t}(\theta) = \mathbb{I}[r_i = 0] \cdot R_{i,t}(\theta) \), where \( r_i \in \{0, 1\} \) is the binary outcome reward. This avoids penalizing correct responses.

\textbf{Group-Level Masking.}  
We further extend masking to the group level: if any response within a sampled group \( \{o_i\}_{i=1}^G \) receives a reward, the KD loss is suppressed for the entire group. Formally, we define \(R'_{i,t}(\theta) = \mathbb{I}[\forall j \in \{1, \dots, G\},\, r_j = 0] \cdot R_{i,t}(\theta)\). This stricter masking criterion ensures that the KD loss is only applied for hard queries when all responses fail.

\textbf{Results.} 
As depicted in Figure~\ref{fig:masking}, we find that response-level masking achieves performance comparable to no masking, but with shorter responses and higher KL divergence. This suggests that selective imitation of low-reward (incorrect) responses enables more efficient learning by avoiding over-regularization on already correct outputs.  
In contrast, group-level masking yields limited improvement, likely due to under-utilization of teacher signals.

\begin{takeawaybox}{Takeaway 3.4 -  Reward-Guided KD}
Response-level reward-guided masking achieves better training and reasoning token efficiency with comparable performance.
\end{takeawaybox}

\begin{table}[H]
\caption{Main results across multiple mathematical reasoning datasets. The best result in each column is highlighted in bold, and the second-best is underlined. AIME24, AIME25, and AMC23 are reported with Avg@16 performance, while others are Pass@1 results.
}
\label{table:main}
\centering
\footnotesize
\begin{tabular}{@{}p{3.3cm}@{}C{1.5cm}@{}C{1.5cm}@{}C{1.5cm}@{}C{1.5cm}@{}C{1.6cm}@{}C{1.6cm}@{}C{1.2cm}@{}}
\toprule
\multirow{2}{*}{\textbf{Model}} & \multirow{2}{*}{\textbf{AIME24}} & \multirow{2}{*}{\textbf{AIME25}} & \textbf{MATH} & \multirow{2}{*}{\textbf{AMC23}} & \textbf{Minerva} & \textbf{Olympiad} & \multirow{2}{*}{\textbf{Avg.}} \\
 &  &  & \textbf{500} &  & \textbf{Math} & \textbf{Bench} &  \\
\midrule
R1-Distill-Qwen-7B & 55.5 & {39.2} & 92.8 & 90.2 & {45.2} & {67.4} & {65.1} \\
Skywork-OR1-Math-7B & {69.8} & {52.3} & {95.8} & {94.1} & {49.3} & {73.5} & {72.5} \\
R1-Distill-Qwen-1.5B & 28.9 & 21.0 & 83.9 & 71.7 & 36.8 & 49.9 & 48.7 \\
\cmidrule{1-8}
DeepScaleR-1.5B-8K & 30.8 & 22.9 & 86.6 & 76.1 & 36.4 & 55.6 & 51.4 \\
\cdashline{1-8}
\addlinespace[1.5pt]
\quad w/ SFT & 33.5 & 24.6 & 87.6 & 76.9 & 31.6 & 54.8 & 51.5 \\
\quad w/ KD-RKL & 41.0 & \underline{29.1} & 89.0 & 80.2 & 37.9 & \underline{59.6} & 56.1 \\
\quad w/ GRPO & 38.3 & 27.1 & 88.6 & 79.5 & 37.5 & 56.3 & 54.6 \\
\quad w/ KDRL & \underline{42.1} & 28.8 & 90.0 & \underline{81.3} & \textbf{39.3} & 59.4 & \underline{56.8} \\
\quad w/ KDRL-Annealing & \textbf{42.9} & \textbf{29.6} & \textbf{90.4} & \textbf{82.2} & \underline{38.2} & \textbf{60.0} & \textbf{57.2} \\
\bottomrule
\end{tabular}
\vspace{-1mm}
\end{table}

\section{Experiments}
\subsection{Experimental Setup}
\label{sec:setting}
\textbf{Training.}  
Following prior studies~\cite{deepscaler2025, skywork-or1-2025}, we adopt a multi-stage training protocol, in which the second stage is the focal point of our study. \textbf{Stage 1 merely establishes a high-efficiency starting point}: we train \textit{R1-Distill-Qwen-1.5B}~\cite{guo2025deepseek} on the DeepScaleR dataset~\cite{deepscaler2025} under an 8K context window to reduce reasoning length and improve subsequent training efficiency, resulting in the backbone \textit{DeepScaleR-1.5B-8K}. 
\textbf{Stage 2 constitutes the core of our investigation:} we perform KDRL using \textit{Skywork-OR1-Math-7B}~\cite{skywork-or1-2025} as the teacher to provide on-policy KD signals. Each training step samples 16 rollouts per prompt with a context length of 20K. The training data is drawn from the \textit{Skywork-OR1-RL-Data} corpus, with difficulty-controlled sampling to ensure stable learning. Further details on data preparation and training configurations are provided in Appendix~\ref{appendix:training}.

\textbf{Baselines.}  
To evaluate the effectiveness of KDRL, we compare it against the following baselines: 
\textit{(i)}~\textbf{SFT} performs SFT by reject-sampling from multiple teacher outputs;  
\textit{(ii)}~\textbf{KD-RKL} applies on-policy distillation via reverse KL;  
\textit{(iii)}~\textbf{GRPO} serves as a pure RL baseline.  
We also include two variants of our method:  
\textit{(iv)}~\textbf{KDRL} with a fixed KL coefficient $\beta$ as 2e-3 using $k2$ estimator, and  
\textit{(v)}~\textbf{KDRL-Annealing}, which anneals the KL coefficient $\beta$ by setting $\beta_{\text{init}}$ to 5e-3, $\delta$ to 5e-5, and $\beta_{\text{min}}$ to 1e-3.
All methods are trained for 280 steps under identical settings for a fair comparison. More details are provided in Appendix~\ref{appendix:baselines}.

\textbf{Evaluation.}  
We evaluate our framework on several widely-used mathematical reasoning benchmarks, including MATH500~\cite{math500}, MinervaMath~\cite{minervamath}, and OlympiadBench~\cite{he-etal-2024-olympiadbench}, as well as competition-level datasets such as AIME24~\cite{aime24}, AIME25~\cite{aime245}, and AMC23~\cite{amc23}.  
For the AIME and AMC datasets, we calculate the average score over 16 decoding runs (Avg@16), while we report Pass@1 accuracy for the remaining datasets. The evaluation criterion is based on the accuracy reward detailed in Appendix~\ref{appendix:training}.
All evaluations are performed under a 32K context window using vLLM~\cite{kwon2023efficient}, with a temperature of 0.6 and top-p set to 1.0.

\subsection{Main Results}

Table~\ref{table:main} presents the performance of all methods across six mathematical reasoning benchmarks. We summarize three key findings below:

(1) Both KDRL variants consistently outperform their standalone KD and RL counterparts. In particular, {KDRL-Annealing} achieves the highest overall accuracy, improving over {SFT}, {GRPO}, {KD-RKL} by 4.7\%, 2.6\%, and 1.1\%, respectively. These results confirm the benefit of integrating teacher guidance and reward feedback within a unified training objective. Compared to {KDRL} with a fixed KL coefficient, {KDRL-Annealing} yields a modest improvement, suggesting that KL annealing enables a better trade-off between imitation and reward-driven learning.

(2) Among the KD and RL baselines, {SFT} performs the worst, falling 3.1\% below the on-policy RL baseline {GRPO}, suggesting that off-policy supervision fails to provide stable and effective guidance. In contrast, {KD-RKL} surpasses {GRPO} by 1.5\%, underscoring the advantage of on-policy distillation. 
These findings are consistent with recent results from the contemporaneous Qwen3 study~\cite{yang2025qwen3technicalreport}.

(3) Additionally, we find that the utilized teacher model, Skywork-OR1-Math-7B, significantly outperforms its backbone R1-Distill-Qwen-7B, demonstrating the strength of our teacher supervision. This aspect will be further discussed in Section~\ref{sec:ablation}. Further, we observe that DeepScaleR-1.5B-8K, trained under an 8K context constraint, improves over its initialization by 2.7\%. This indicates that limiting output length during the first-stage training does not impair reasoning performance, and instead improves training efficiency in subsequent stages.

\begin{figure*}[t]
    \centering
    \parbox[b]{0.48\textwidth}{
        \centering
        \includegraphics[width=\linewidth]{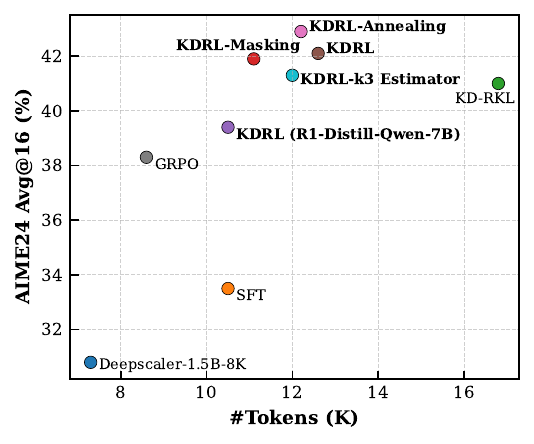}
    }
    \hfill
    \parbox[b]{0.48\textwidth}{
        \centering
        \includegraphics[width=\linewidth]{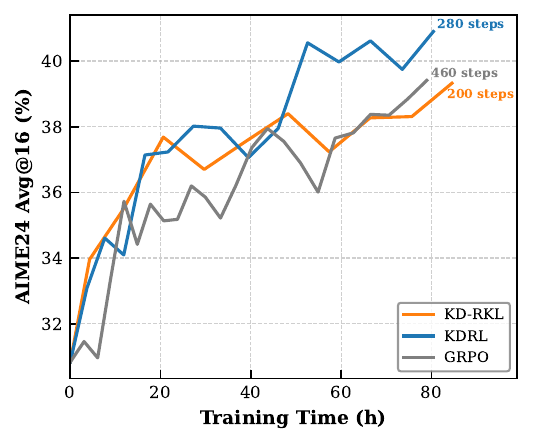}
    }
    \vspace{-2mm}
    \caption{We evaluate the trade-offs between efficiency and performance. \textbf{Left:} Reasoning token efficiency versus AIME24 accuracy. \textbf{Right:} AIME24 accuracy over cumulative training time.}
    \vspace{-2mm}
    \label{fig:efficiency}
\end{figure*}

\subsection{Efficiency Analysis}
\label{sec:efficiency} 

\textbf{Reasoning Token Efficiency.}  
To evaluate the trade-off between accuracy and reasoning efficiency, we analyze model performance and average output length on the AIME24 dataset, as shown in Figure~\ref{fig:efficiency} (left). We observe that all KD and RL methods increase output length as they improve performance. Among them, the proposed {KDRL} and {KDRL-Annealing} achieve the best balance, outperforming the baselines with moderate token efficiency. In contrast, {GRPO} yields a modest increase in length but shows limited gains in accuracy. {KD-RKL} performs well but produces significantly longer outputs, which cost more than 6K additional tokens compared to {KDRL}, resulting in inefficient inference. Additionally, while {SFT} leads to longer outputs than {GRPO}, it performs the worst among all variants, indicating that off-policy guidance struggles to extend the effective thoughts.

\noindent\textbf{Training Efficiency.}
To ensure a fair comparison of training efficiency, we align the total training time across methods, which naturally varies due to differences in output length and teacher model usage. As shown in Figure~\ref{fig:efficiency} (right), using KDRL as the reference, 280 training steps correspond roughly to 200 steps of {KD-RKL} and 460 steps of GRPO, each totaling around 80 hours. Specifically, {GRPO} trains fastest due to shorter outputs and the absence of teacher inference, while {KD-RKL} is slowed by significantly longer sequences and distillation overhead. {KDRL} lies in between in terms of efficiency, yet consistently outperforms both methods under equal training budgets. Moreover, we find that {GRPO} achieves performance comparable to the on-policy distillation. These results suggest that imitation and exploration may offer similar potential individually in our setting, but our unified {KDRL} framework leverages both more effectively to achieve superior performance.
\subsection{Ablation Study}
\label{sec:ablation} 

To further assess the contribution of individual components, we conduct an ablation study on the AIME24 dataset. We consider three variants: (1) KDRL-$k3$ Estimator, which replaces the $k2$ estimator in the KL approximation with $k3$; (2) KDRL (R1-Distill-Qwen-7B), which substitutes the teacher model with a weaker alternative; and (3) KDRL-Masking, which applies response-level masking as described in Section~\ref{sec:masking}. Results are presented in Figure~\ref{fig:efficiency} (left). We find that KDRL-k3 Estimator leads to shorter outputs but results in a slight drop in accuracy, confirming the superiority of the $k2$ estimator in our framework. Nevertheless, it still outperforms all baselines, demonstrating the robustness of the KDRL objective. Using R1-Distill-Qwen-7B as the teacher provides slight gains over GRPO (+1.1\%), but falls short of KD-RKL and KDRL with the stronger Skywork-OR1-Math-7B. This may be attributed to both the substantially lower accuracy of the weaker teacher and its lack of RL-based tuning, which may limit its utility for reward-aligned guidance. Finally, KDRL-Masking reduces output length by over 1.4K tokens (more than 10\%) while maintaining comparable accuracy, suggesting that selective distillation on low-reward responses provides an effective trade-off between inference efficiency and performance.

\subsection{KDRL for R1-Zero-Like Training}

The advent of the DeepSeek-R1-Zero model~\cite{guo2025deepseek} demonstrates that applying rule-based RL directly to a pre-trained language model can significantly enhance reasoning by promoting longer and more deliberate thinking. Subsequent studies have adopted PPO or GRPO for R1-Zero-like training~\cite{liu2025understanding,yu2025dapo,hu2025openreasonerzeroopensourceapproach}. To evaluate whether KDRL can similarly improve such “zero-RL” training by integrating the strengths of KD and RL, we compare the effectiveness of {KDRL-Annealing} with {KD-RKL} and {GRPO}. We describe the experimental setup and findings as follows.

\textbf{Settings.} We adopt \textit{Qwen2.5-3B}~\cite{yang2024qwen25} as the policy model for training. 
To implement KDRL, we use \textit{Qwen2.5-7B}, a larger variant from the same model family, and apply GRPO training to construct the teacher model \textit{Qwen2.5-7B-GRPO}, ensuring similar output distributions as the student. The training corpus is consistent with that described in Section~\ref{sec:setting}, with difficulty rebalancing performed to align with the capacity of the base model and ensure stable training. 
We follow the main experimental setup with several key adjustments (e.g., context length, KL coefficient) to better accommodate the characteristics of zero-RL. Additional training details are provided in Appendix~\ref{appendix:zero}. For evaluation, we use the same datasets and criteria as in Section~\ref{sec:setting}, with the context length adapted to 8K.

\begin{wrapfigure}{r}{0.5\textwidth}
  \centering
  \vspace{-10pt}
  \includegraphics[width=1\linewidth]{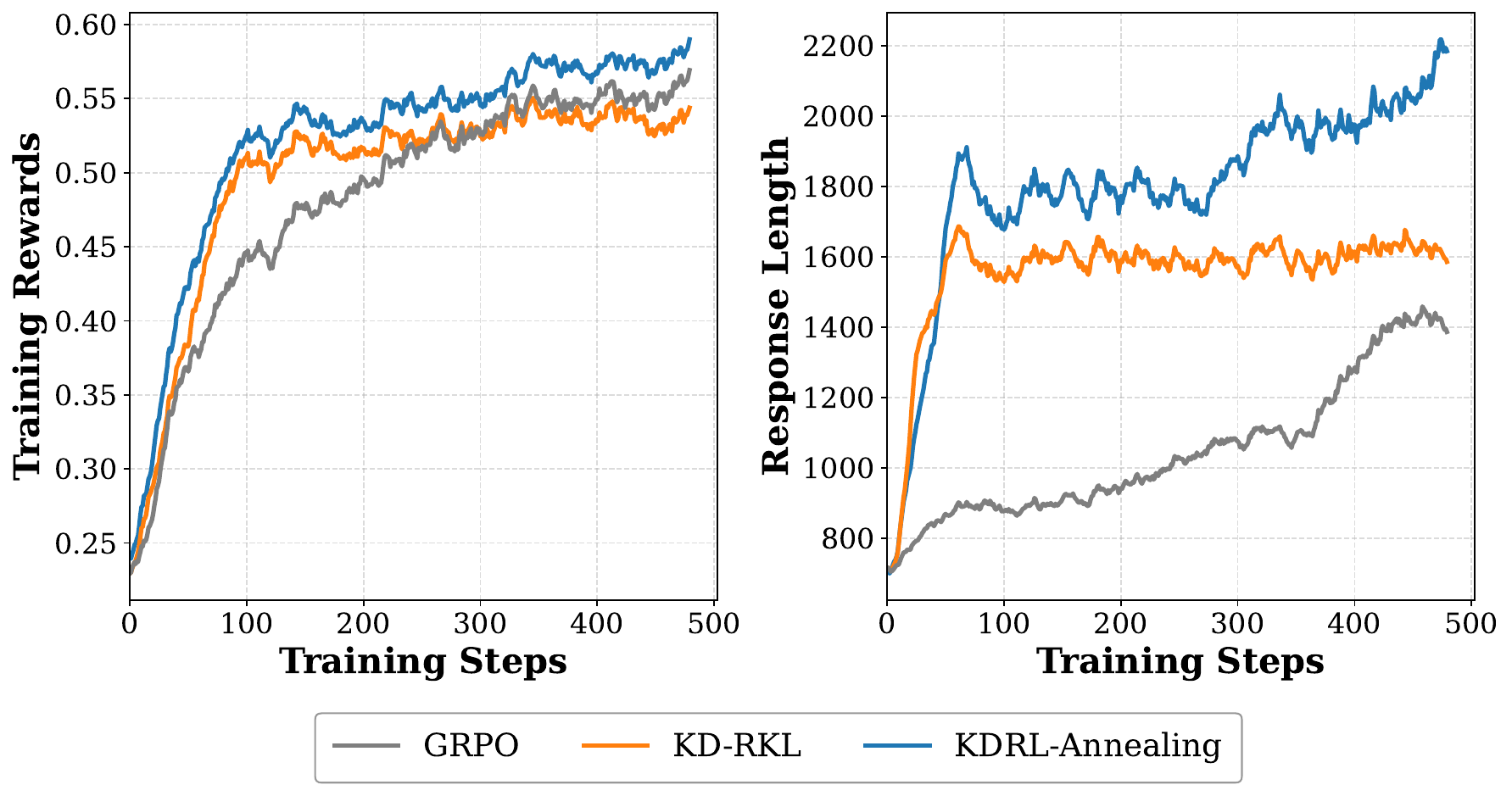}
  \vspace{-10pt}
  \caption{Training rewards and response length dynamics during R1-Zero-like training.}
  \label{fig:zero_train}
  \vspace{-5pt}
\end{wrapfigure}

\textbf{Training Dynamics.}  
Figure~\ref{fig:zero_train} depicts the training curves of R1-Zero-like training. We observe that KDRL-Annealing consistently achieves higher training rewards than both baselines throughout training, indicating more effective policy optimization. In the early phase, KD-RKL yields slightly better rewards than GRPO; however, it saturates quickly, while GRPO gradually catches up and eventually surpasses KD-RKL, highlighting the stronger long-term exploration capability of direct reward optimization. 
In terms of response length, GRPO shows steady incremental growth, while KD-RKL leads to a rapid initial increase that soon plateaus. In contrast, KDRL-Annealing benefits from its balanced integration of teacher supervision and self-exploration, facilitating both faster early-stage growth and continued expansion in later stages.

\textbf{Evaluation.} As shown in Table~\ref{table:zero}, KDRL-Annealing achieves the best performance on all datasets, surpassing GRPO and KD-RKL by 1.2\% and 1.7\%, respectively. These results validate the effectiveness of unifying KD and RL with KL scheduling. 
Relative to the base model, all three post-training strategies yield substantial improvements, demonstrating the clear benefit of on-policy optimization under the R1-Zero-like setting. 
Between the two baselines, GRPO slightly outperforms KD-RKL (31.3\% vs. 30.8\%), suggesting that the upper bound of on-policy distillation is not always higher than RL.
By balancing teacher supervision with reward-guided exploration, KDRL-Annealing achieves consistent gains across all benchmarks.

\begin{table}[H]
\caption{Performance of R1-Zero-like training using Qwen2.5-3B. AIME24, AIME25, and AMC23
are reported with Avg@16 performance, while others are Pass@1 results.}
\label{table:zero}
\centering
\footnotesize
\begin{tabular}{@{}p{3.3cm}@{}C{1.5cm}@{}C{1.5cm}@{}C{1.5cm}@{}C{1.5cm}@{}C{1.6cm}@{}C{1.6cm}@{}C{1.2cm}@{}}
\toprule
\multirow{2}{*}{\textbf{Model}} & \multirow{2}{*}{\textbf{AIME24}} & \multirow{2}{*}{\textbf{AIME25}} & \textbf{MATH} & \multirow{2}{*}{\textbf{AMC23}} & \textbf{Minerva} & \textbf{Olympiad} & \multirow{2}{*}{\textbf{Avg}}\\
 &  &  & \textbf{500} &  & \textbf{Math} & \textbf{Bench} &  \\
\midrule
Qwen2.5-3B              & 3.8  & 0.6 & 55.6 & 25.6 & 19.5 & 22.8 & 21.3 \\
\cdashline{1-8}
\addlinespace[1.5pt]
\quad w/ KD-RKL         & 7.9  & \underline{5.2} & 69.6 & \underline{43.1} & 26.8 & 31.9 & 30.8 \\
\quad w/ GRPO           & \underline{9.0}  & 4.6 & \underline{70.6} & 43.0 & \underline{27.6} & \underline{32.9} & \underline{31.3} \\
\quad w/ KDRL-Annealing & \textbf{10.4} & \textbf{5.6} & \textbf{71.4} & \textbf{44.5} & \textbf{29.0} & \textbf{34.2} & \textbf{32.5} \\
\bottomrule
\end{tabular}
\vspace{-2mm}
\end{table}

\section{Related Work}
\paragraph{Reinforcement Learning for Reasoning LLMs.}
The recent success of Kimi K1.5~\cite{team2025kimi} and DeepSeek-R1~\cite{guo2025deepseek} has demonstrated the feasibility of training long-chain-of-thought (long-CoT) reasoning LLMs using reinforcement learning (RL). 
Building on this foundation, a range of RL-based approaches have been proposed to enhance reasoning capabilities, which can be broadly grouped into three categories:
(i) applying RL directly to pretrained models, as exemplified by R1-Zero-like training~\cite{xie2025logicrlunleashingllmreasoning,hu2025openreasonerzeroopensourceapproach,zeng2025simplerlzooinvestigatingtamingzero};
(ii) enhancing distilled models to unlock the reasoning boundary of smaller LLMs~\cite{deepscaler2025,skywork-or1-2025,wen2025lightr1curriculumsftdpo,mei2025real};
(iii) developing model-agnostic strategy improvements that are decoupled from specific backbones~\cite{yu2025dapo,liu2025understanding,zhang2025grpoleaddifficultyawarereinforcementlearning,yue2025vapoefficientreliablereinforcement}.
Our proposed method falls into the third category, with experiments demonstrating its effectiveness in both Zero-RL and RL-from-distilled settings. In contrast to prior work, it incorporates teacher supervision during reinforcement fine-tuning to facilitate more effective reasoning.

\paragraph{Knowledge Distillation.}
Knowledge distillation (KD) for LLMs can be broadly categorized into two paradigms based on whether the student learns from teacher-guided outputs (off-policy) or its own samples (on-policy). 
(i) \textit{Standard KD} typically relies on teacher-sampled distributions and can be applied either at the sequence level~\cite{hsieh-etal-2023-distilling,guo2025deepseek} or the token-logit level~\cite{liu2024ddk,gu2025miniplm}. 
(ii) \textit{On-policy KD} optimizes based on student-generated samples and corresponding teacher logits, which helps mitigate exposure bias and supports reinforcement-style training~\cite{gu2024minillm,ko2024distillm,yang2025qwen3technicalreport}.
Most prior work on distilling reasoning LLMs has focused on standard SFT-style off-policy approaches~\cite{guo2025deepseek,ye2025limoreasoning,muennighoff2025s1simpletesttimescaling,tian2025deepdistillenhancingllmreasoning}, while Qwen3~\cite{yang2025qwen3technicalreport} explores on-policy distillation and demonstrates its promising potential. 
~\cite{yang2025qwen3technicalreport}
Our proposed method, KDRL, builds upon the on-policy paradigm by integrating RL with KD-RKL, enabling a unified objective that benefits from both imitation and exploration.

\paragraph{Combination of RL and KD.}
Previous work has explored integrating reinforcement learning with knowledge distillation to jointly leverage reward signals and teacher supervision. One line of studies incorporates teacher guidance into preference optimization by modifying the DPO objective~\cite{liu2024ddk, pan2025predpoimprovingdatautilization}. In addition, GKD~\cite{agarwal2024onpolicy} proposes a generalized on-policy distillation framework and first explores unifying RL and KD for text summarization. More recently, LUFFY~\cite{luffy} combines off-policy trajectories with policy shaping to align SFT and RL for training reasoning LLMs.
The proposed KDRL shares the high-level goal of GKD but introduces two key innovations. First, it systematically investigates the integration of on-policy distillation with rule-based RL for reasoning LLMs, providing a detailed analysis of both training dynamics and performance. This marks one of the first steps toward a unified post-training framework for long-CoT models. Second, KDRL incorporates several targeted techniques, including KL approximation with annealing to adaptively guide learning, and reward-guided masking to improve training efficiency. We believe KDRL offers practical guidance and provides valuable insights into post-training strategies for the era of deep reasoning.

\section{Conclusion}

In this study, we introduce KDRL, a unified post-training framework that enhances the reasoning capabilities of LLMs by integrating teacher guidance (knowledge distillation) and self-improvement (reinforcement learning) via policy gradients under an on-policy optimization scheme.
We perform comprehensive analyses of how different KL approximation methods, KL coefficients, and reward-guided distillation influence training dynamics, model performance, and token efficiency. Experimental results across several reasoning benchmarks show that KDRL not only surpasses GRPO and various KD baselines in performance, but also achieves greater training and reasoning token efficiency. These outcomes suggest that combining KD with RL offers a powerful and resource-efficient approach to scaling reasoning LLMs post-training.

\clearpage

\bibliographystyle{unsrt}
\bibliography{citation}

\clearpage
\newpage

\appendix
\section{Mathematical Derivations}
\label{appendix:math}

\subsection{Reinforcement Learning}
We derive the gradient of REINFORCE and our adopted GRPO algorithm below.

\begin{align}
\mathcal{J}_{\textrm{REINFORCE}}(\theta)&=\mathbb{E}_{q\sim Q, o\sim \pi_{\theta}(\cdot|q)}[r(o,q)], \\
\nabla_\theta \mathcal{J}_{\textrm{REINFORCE}}(\theta)&=\mathbb{E}_{q\sim Q}\bigg[\nabla_\theta\sum_{o} \pi_{\theta}(o|q) r(o,q)\bigg] \nonumber \\
&=\mathbb{E}_{q\sim Q}\bigg[\sum_{o} r(o,q)\nabla_\theta\pi_{\theta}(o|q) \bigg] \nonumber  \\
&=\mathbb{E}_{q\sim Q}\bigg[\sum_{o} r(o,q)\pi_{\theta}(o|q)\nabla_\theta \log \pi_{\theta}(o|q) \bigg] \nonumber \\
&=\mathbb{E}_{q\sim Q, o\sim \pi_{\theta}(\cdot|q)}\bigg[ r(o,q)\nabla_\theta \log \pi_{\theta}(o|q) \bigg].
\end{align}

\begin{align}
    \mathcal{J}_{\text{GRPO}}(\theta) &= 
    \mathbb{E}_{q\sim Q ,\, \{o_i\}_{i=1}^G\sim \pi_{\theta_{old}}(\cdot|q)}  
    \bigg[ \frac{1}{\sum_{i=1}^G|o_i|} \sum_{i=1}^{G} \sum_{t=1}^{|o_i|} \rho_{i,t}(\theta) 
    \hat{A}_{i} \bigg], \\
    \nabla_\theta \mathcal{J}_{\text{GRPO}}(\theta) &= 
    \mathbb{E}_{q\sim Q ,\, \{o_i\}_{i=1}^G\sim \pi_{\theta_{old}}(\cdot|q)}  
    \bigg[ \frac{1}{\sum_{i=1}^G|o_i|} \sum_{i=1}^{G} \sum_{t=1}^{|o_i|} 
    \hat{A}_{i} \nabla_\theta  \rho_{i,t}(\theta) \bigg]  \nonumber \\
    &= 
    \mathbb{E}_{q\sim Q ,\, \{o_i\}_{i=1}^G\sim \pi_{\theta_{old}}(\cdot|q)}  
    \bigg[ \frac{1}{\sum_{i=1}^G|o_i|} \sum_{i=1}^{G} \sum_{t=1}^{|o_i|} 
    \hat{A}_{i} \frac{\nabla_\theta \pi_\theta(o_{i,t} | q, o_{i,<t})}{\pi_{\theta_{old}}(o_{i,t} | q, o_{i,<t})}  \bigg]  \nonumber \\
    &= 
    \mathbb{E}_{q\sim Q ,\, \{o_i\}_{i=1}^G\sim \pi_{\theta_{old}}(\cdot|q)}  
    \bigg[ \frac{1}{\sum_{i=1}^G|o_i|} \sum_{i=1}^{G} \sum_{t=1}^{|o_i|} 
    \rho_{i,t}(\theta)\hat{A}_{i}    \nabla_\theta \log \pi_\theta(o_{i,t} | q, o_{i,<t})\bigg].
\end{align}

\subsection{Knowledge Distillation}
We show that minimizing the forward KL divergence between the teacher $\pi_T$ and the student $\pi_\theta$ is equivalent to maximizing the log-likelihood on data generated by the teacher $\pi_T$ (aka SFT):
\begin{align}
\arg \min_\theta \mathbb{D}_{\text{KL}}(\pi_T||\pi_\theta)
&=\arg \min_\theta \mathbb{E}_{q\sim Q, o\sim \pi_T(\cdot|q)}\bigg[\log\frac{\pi_T(o|q)}{\pi_\theta(o|q)}\bigg] \nonumber \\
&=\arg \max_\theta \mathbb{E}_{q\sim Q, o\sim \pi_T(\cdot|q)}[\log\pi_\theta(o|q)-\log \pi_T(o|q)] \nonumber \\
&=\arg\max_\theta \;\Bigl(\mathbb{E}_{q\sim Q,o\sim\pi_T(\cdot|q)}[\log\pi_\theta(o|q)]-\underbrace{\mathbb{E}_{q\sim Q,o\sim\pi_T(\cdot|q)}[\log\pi_T(o|q)]}_{\text{constant w.r.t.\ }\theta}\Bigr) \nonumber \\
&=\arg \max_\theta \mathbb{E}_{q\sim Q, o\sim \pi_T(\cdot|q)}[\log\pi_\theta(o|q)] \nonumber \\
&=\arg \max_\theta \mathcal{J}_{\text{SFT}}(\theta).
\end{align}

In practice, the SFT objective is normalized by sequence length:

\begin{align}
\mathcal{J}_{\text{SFT}}(\theta)&=\mathbb{E}_{q\sim Q, o\sim \pi_T(\cdot|q)}\bigg[\frac{1}{|o|}\sum_{t=1}^{|o|}\log\pi_\theta(o_t|q,o_{<t})\bigg],\\
\nabla_\theta \mathcal{J}_{\text{SFT}}(\theta)&=\mathbb{E}_{q\sim Q, o\sim \pi_T(\cdot|q)}\bigg[\frac{1}{|o|}\sum_{t=1}^{|o|}\nabla_\theta\log\pi_\theta(o_t|q,o_{<t})\bigg].
\end{align}

Different from SFT, distilling knowledge by minimizing the reverse KL divergence requires sampling from the current policy $\pi_\theta$.
When adopting similar techniques as GRPO but simplifying the notation of group-sampling in Eq. \ref{eq:RKL}, the RKL objective and its gradient are:

\begin{align}
\mathcal{J}_{\text{RKL}}(\theta) &= \mathbb{E}_{q \sim Q, o \sim \pi_{\theta_{old}}(\cdot|q)} \bigg[\frac{1}{|o|} \sum_{t=1}^{|o|} \rho_{t}(\theta) R_t(\theta) \bigg], \\
\nabla_\theta\mathcal{J}_{\text{RKL}}(\theta) &= \mathbb{E}_{q \sim Q, o \sim \pi_{\theta_{old}}(\cdot|q)} \bigg[\frac{1}{|o|} \sum_{t=1}^{|o|}  \big(R_t(\theta)\nabla_\theta \rho_{t}(\theta)+\rho_{t}(\theta)\nabla_\theta R_t(\theta)\big)\bigg] \nonumber \\
&= \mathbb{E}_{q \sim Q, o \sim \pi_{\theta_{old}}(\cdot|q)} \bigg[\frac{1}{|o|} \sum_{t=1}^{|o|} \rho_{t}(\theta)R_t(\theta) \nabla_\theta\log\pi_\theta(o_{t} | q, o_{<t})\bigg] \nonumber \\
&\quad - \mathbb{E}_{q \sim Q, o \sim \pi_{\theta_{old}}(\cdot|q)} \bigg[\frac{1}{|o|} \sum_{t=1}^{|o|}\frac{\nabla_\theta \pi_\theta(o_t|q,o_{<t})}{\pi_{\theta_{old}}(o_t|q,o_{<t})}\bigg] \nonumber \\
&= \mathbb{E}_{q \sim Q, o \sim \pi_{\theta_{old}}(\cdot|q)} \bigg[\frac{1}{|o|} \sum_{t=1}^{|o|} \rho_{t}(\theta)R_t(\theta) \nabla_\theta\log\pi_\theta(o_{t} | q, o_{<t})\bigg] \nonumber \\
&\quad - \nabla_\theta\mathbb{E}_{q \sim Q, o \sim \pi_{\theta_{old}}(\cdot|q)} \bigg[\frac{1}{|o|} \sum_{t=1}^{|o|}\frac{\pi_\theta(o_t|q,o_{<t})}{\pi_{\theta_{old}}(o_t|q,o_{<t})}\bigg] \nonumber \\
&= \mathbb{E}_{q \sim Q, o \sim \pi_{\theta_{old}}(\cdot|q)} \bigg[\frac{1}{|o|} \sum_{t=1}^{|o|} \rho_{t}(\theta)R_t(\theta) \nabla_\theta\log\pi_\theta(o_{t} | q, o_{<t})\bigg] \nonumber \\
&\quad - \nabla_\theta\mathbb{E}_{q \sim Q, o \sim \pi_{\theta}(\cdot|q)} \bigg[\frac{1}{|o|} \sum_{t=1}^{|o|}1\bigg] \nonumber \\
&= \mathbb{E}_{q \sim Q, o \sim \pi_{\theta_{old}}(\cdot|q)} \bigg[\frac{1}{|o|} \sum_{t=1}^{|o|} \rho_{t}(\theta)R_t(\theta) \nabla_\theta\log\pi_\theta(o_{t} | q, o_{<t})\bigg],
\end{align}
where $\rho_t(\theta)=\frac{\pi_\theta(o_t|q,o_{<t})}{\pi_{\theta_{old}}(o_t|q,o_{<t})}=1$ and $R_t(\theta)=\log\frac{\pi_T(o_t|q,o_{<t})}{\pi_{\theta}(o_t|q,o_{<t})}$.

\subsection{KL Approximation}
We present several approximations of the reverse KL divergence (RKL) and their corresponding token-level gradients.  
For notational simplicity, normalization by sequence length is omitted.

\textbf{Direct KL estimation.}  
We begin with the standard Monte Carlo estimate of the RKL, which serves as a reference for the subsequent approximations:
\begin{align}
    \mathbb{D}_{\mathrm{KL}}(\pi_\theta \| \pi_T) &= \mathbb{E}_{o_t \sim \pi_{\theta}(\cdot|q,o_{<t})}[- R_t(\theta) ],\\
    \nabla_\theta\mathbb{D}_{\mathrm{KL}}(\pi_\theta \| \pi_T) &= \nabla_\theta\sum_{o_t}-\pi_{\theta}(o_t|q,o_{<t})R_t(\theta) \nonumber \\
    &= \sum_{o_t}\big(-\pi_{\theta}(o_t|q,o_{<t})R_t(\theta)+\pi_{\theta}(o_t|q,o_{<t})\big)\nabla_\theta\log \pi_{\theta}(o_t|q,o_{<t})  \nonumber \\
    &= \mathbb{E}_{o_t \sim \pi_{\theta}(\cdot|q,o_{<t})}[ -R_t(\theta)\nabla_\theta\log \pi_{\theta}(o_t|q,o_{<t})]+\nabla_\theta \sum_{o_t}\pi_{\theta}(o_t|q,o_{<t}) \nonumber \\
    &= \mathbb{E}_{o_t \sim \pi_{\theta}(\cdot|q,o_{<t})}[ -R_t(\theta)\nabla_\theta\log \pi_{\theta}(o_t|q,o_{<t})]+\nabla_\theta 1 \nonumber \\
    &= \mathbb{E}_{o_t \sim \pi_{\theta}(\cdot|q,o_{<t})}[ -R_t(\theta)\nabla_\theta\log \pi_{\theta}(o_t|q,o_{<t})].
\end{align}

\textbf{$k2$ approximation.} 
The $k2$ estimator provides an unbiased estimate of the RKL gradient:
\begin{align}
    \mathbb{D}_{\mathrm{KL}}^{k2}(\pi_\theta \| \pi_T) &= \mathbb{E}_{o_t \sim \pi_{\theta_{old}}(\cdot|q,o_{<t})}\bigg[\frac{1}{2} R_t(\theta)^2 \bigg],\\
    \nabla_\theta\mathbb{D}_{\mathrm{KL}}^{k2}(\pi_\theta \| \pi_T) &= \mathbb{E}_{o_t \sim \pi_{\theta_{old}}(\cdot|q,o_{<t})}\bigg[R_t(\theta)\nabla_\theta R_t(\theta) \bigg] \nonumber \\\
    &= \mathbb{E}_{o_t \sim \pi_{\theta_{old}}(\cdot|q,o_{<t})}\bigg[-R_t(\theta)\nabla_\theta \log\pi_\theta(o_t|q,o_{<t}) \bigg].
\end{align}

\textbf{$k3$ approximation.}
While the $k3$ estimator provides an unbiased estimate of the scalar divergence, its gradient is biased and introduces asymmetric teacher signals:
\begin{align}
    \mathbb{D}_{\mathrm{KL}}^{k3}(\pi_\theta \| \pi_T) &= \mathbb{E}_{o_t \sim \pi_{\theta_{old}}(\cdot|q,o_{<t})}\left[e^{R_t(\theta)} - R_t(\theta) - 1 \right] \nonumber \\\
    &= \mathbb{E}_{o_t \sim \pi_{\theta_{old}}(\cdot|q,o_{<t})}\left[- R_t(\theta) \right]+\sum_{o_t}\left[\pi_{\theta_{old}}(o_t|q,o_{<t})\cdot \frac{\pi_T(o_t|q,o_{<t})}{\pi_{\theta}(o_t|q,o_{<t})}\right]- 1 \nonumber \\\
    &= \mathbb{E}_{o_t \sim \pi_{\theta_{old}}(\cdot|q,o_{<t})}\left[- R_t(\theta) \right], \\
    \nabla_\theta \mathbb{D}_{\mathrm{KL}}^{k3}(\pi_\theta \| \pi_T) &= \mathbb{E}_{o_t \sim \pi_{\theta_{old}}(\cdot|q,o_{<t})}\left[(e^{R_t(\theta)}-1)\nabla_\theta R_t(\theta) \right] \nonumber \\\
    &= \mathbb{E}_{o_t \sim \pi_{\theta_{old}}(\cdot|q,o_{<t})}\left[-(e^{R_t(\theta)}-1)\nabla_\theta \log \pi_\theta(o_t|q,o_{<t}) \right].
\end{align}

\textbf{Top-$K$ approximation.}
While supervised learning typically computes KL divergence over the entire vocabulary $V$, RL training often relies on Monte Carlo estimation over sampled tokens to reduce computational cost. However, this limits the utilization of teacher signals and may lead to suboptimal guidance. A practical alternative is the Top-$K$ approximation: 
\begin{align}
\mathbb{D}_{\mathrm{KL}}^{\textit{Top-K}} (\pi_\theta \| \pi_T) &= \mathbb{E}_{o_t \sim \pi_{\theta_{old}}(\cdot|q,o_{<t})} \bigg[  \sum_{v \in V} \pi_\theta(v|q,o_{<t}) \log \frac{\pi_\theta(v|q,o_{<t})}{\pi_T'(v|q,o_{<t})} \bigg],\\
\nabla_\theta\mathbb{D}_{\mathrm{KL}}^{\textit{Top-K}} (\pi_\theta \| \pi_T) &= \mathbb{E}_{o_t \sim \pi_{\theta_{old}}(\cdot|q,o_{<t})} \bigg[  \sum_{v \in V}  (\log \frac{\pi_\theta(v|q,o_{<t})}{\pi_T'(v|q,o_{<t})}+1)\nabla_\theta 
 \pi_\theta(v|q,o_{<t})\bigg] \nonumber \\
 &= \mathbb{E}_{o_t \sim \pi_{\theta_{old}}(\cdot|q,o_{<t})} \bigg[  \sum_{v \in V}\pi_\theta(v|q,o_{<t})  \log \frac{\pi_\theta(v|q,o_{<t})}{\pi_T'(v|q,o_{<t})}\nabla_\theta 
 \log\pi_\theta(v|q,o_{<t})\bigg].
\end{align}
Here, we retain the logits of the top-$K$ tokens in $\pi_T(v|q,o_{<t})$, set all others to $-\infty$, and apply the softmax function to obtain the normalized approximation $\pi_T'(v|q,o_{<t})$.

\subsection{KDRL}
We show that the gradient of $\mathcal{J}_{\text{KDRL}}(\theta)$ is the combination of $\mathcal{J}_{\text{GRPO}}(\theta)$ and $\mathcal{J}_{\text{RKL}}(\theta)$:
\begin{align}
    \mathcal{J}_{\text{KDRL}}(\theta)
    &= \mathcal{J}_{\text{GRPO}}(\theta) 
    - \beta \mathbb{D}_{\mathrm{KL}}^{k2}(\pi_{\theta}\|\pi_{T})  \nonumber \\
    &=\mathbb{E}_{q\sim Q, \{o_i\}_{i=1}^G\sim \pi_{\theta_{old}}(\cdot|q)}\bigg[
    \frac{1}{\sum_{i=1}^G |o_i|} \sum_{i=1}^{G}\sum_{t=1}^{|o_i|}
    \Big(\rho_{i,t}(\theta)\hat{A}_i-\beta\frac{R^2_{i,t}(\theta)}{2}\Big)\bigg], 
    \\
    \nabla_\theta\mathcal{J}_{\text{KDRL}}(\theta)
    &=\mathbb{E}_{q\sim Q, \{o_i\}_{i=1}^G\sim \pi_{\theta_{old}}(\cdot|q)}\bigg[
    \frac{1}{\sum_{i=1}^G |o_i|} \sum_{i=1}^{G}\sum_{t=1}^{|o_i|}
    \Big(\hat{A}_i+\beta R_{i,t}(\theta)\Big)\nabla_\theta \log \pi_\theta(o_{i,t} | q, o_{i,<t})\bigg]  \nonumber \\
    &=\nabla_\theta \mathcal{J}_{\text{GRPO}}(\theta) +  \beta\cdot\nabla_\theta \mathcal{J}_{\text{RKL}}(\theta). 
\end{align}

\section{Additional Experimental Settings}
\label{appendix:setting}
\subsection{Training Details}
\label{appendix:training}

\subsubsection{Traning Dataset}

\textbf{Training Data for Stage 1 (DeepScaleR-1.5B-8K).}
In the first stage of training for obtaining DeepScaleR-1.5B-8K, we directly reused the open-source dataset released by DeepScaleR~\cite{deepscaler2025}, which contains 40K math question–answer pairs compiled from sources such as AIME, AMC, Omni-Math~\cite{gao2025omnimath}, and so on.

\textbf{Training Data for Stage 2 (KDRL).}
For our main experiments based on DeepScaleR-1.5B-8K, we primarily used the Skywork-OR1 dataset~\cite{skywork-or1-2025}, which comprises a large collection of math problems filtered for verifiability, correctness, and challenge. The dataset further incorporates model-aware difficulty estimation and quality assessments, resulting in 105K high-quality problems.
To better align the data distribution with our training objectives, we apply additional filtering based on difficulty estimation labels. First, we remove all problems with a pass@16 score of at least 15 using the R1-Distill-Qwen-1.5B model. Given that DeepScaleR-1.5B-8K exhibits comparable performance, this step effectively removes trivial samples and improves training efficiency.
We also exclude samples deemed excessively difficult, defined as those answered incorrectly by all three R1-Distill-Qwen models at 1.5B, 7B, and 32B scales. To further balance difficulty, we constrain the proportion of unsolved examples (problems that the 1.5B model fails to solve) to 10\%. These harder cases are retained under the assumption that teacher supervision may provide informative optimization signals during training.
Finally, we exclude problems whose prompts exceed 1024 tokens in length. After all preprocessing steps, the resulting dataset contains \textbf{61K} math problems.

\subsubsection{Implementation Details} 
\label{appendix:Implematation}

\textbf{Reward Function.}
For KDRL training, we adopt a rule-based binary reward function similar to that used in Skywork-OR1~\cite{skywork-or1-2025}. The reward is composed of two components: a format reward and an accuracy reward. A response receives a final reward of 1 only if both components are satisfied, defined as: $r = r_{\text{format}} \cdot r_{\text{acc}}$. 
This reward design encourages the model to produce outputs that are both well-structured and verifiably correct.
The format reward \(r_{\text{format}}\) checks whether the model output follows the expected structure (\texttt{<think>}...\texttt{</think>} pattern). The accuracy reward \(r_{\text{acc}}\) verifies whether the extracted answer matches the reference answer using multi-round verification with \texttt{math-verify} 0.6.0~\cite{Kydlicek_Math-Verify_Math_Verification}. Specifically, if a direct match is not found, the gold answer is wrapped with \texttt{\textbackslash boxed\{\}} and verified again to account for formatting variations.

\textbf{Training Framework and Parameters.} 
We conduct all KDRL and baseline training using the VeRL framework~\cite{sheng2024hybridflow}. Both the Stage 1 training and the KDRL training use a similar set of core hyper-parameters, except that KDRL introduces additional parameters associated with the proposed strategies, as discussed below.
We use a consistent prompting format across the main experiments: \texttt{"\{question\} Let's think step by step and output the final answer within \textbackslash boxed\{\}."} This prompt is wrapped using the chat template of the R1-Distill-Qwen model family.
The key hyper-parameters used for training are summarized in Table~\ref{tab:hyperparameters}.

\begin{table}[t]
\centering
\caption{Training hyper-parameters used in the two-stage pipeline. GRPO and KD-RKL baselines share the same settings as KDRL, except for KL-related differences.}
\label{tab:hyperparameters}
\begin{tabular}{lcc}
\toprule
\textbf{Hyper-parameter} & \textbf{Stage 1: DeepScaleR-1.5B-8K} & \textbf{Stage 2: KDRL} \\
\midrule
Train batch size & 64 & 64 \\
PPO mini-batch size & 64 & 64 \\
Number of rollouts & 16 & 16 \\
Learning rate & 1e-6 & 1e-6 \\
Sampling temperature & 0.6 & 0.6 \\
Max prompt length & 1,024 & 1,024 \\
Max response length & 8,192 & 20,480 \\
Training steps & 500 & 280 \\
Entropy coefficient & 1e-3 & 1e-3 \\
KL coefficient $\beta$ & 1e-3 & 2e-3 (or annealed: 5e-3 $\rightarrow$ 1e-3) \\
KL loss type & \texttt{low\_var\_kl} & \texttt{mse} \\
\bottomrule
\end{tabular}
\vspace{-2mm}
\end{table}

\subsection{Baseline Descriptions}
\label{appendix:baselines}
We detail the implementation of each baseline used in our experiments below.

\textbf{(i) SFT} performs supervised fine-tuning using teacher-generated outputs. Candidate responses are sampled eight times under a 20K token context window, and only those matching the ground-truth answer are retained via reject sampling. The model is trained with a global batch size of 128 and a learning rate of 2e-7. Empirically, we find that relatively small learning rates are required to ensure stable optimization in this setting.

\textbf{(ii) KD-RKL} minimizes the RKL between the student and teacher distributions, where a fixed KL coefficient of 2e-3 is used. Other hyper-parameters are aligned with KDRL settings.

\textbf{(iii) GRPO} is a pure RL baseline optimized using the objective defined in Eq.~\ref{eq:GRPO_NEW}. This method does not incorporate teacher supervision, which results in faster training but lacks imitation signals. All hyper-parameters except KL computation are kept consistent with those in KDRL training.

\subsection{Details of R1-Zero-like Training}
\label{appendix:zero}

\textbf{Training Data.}
We also use the Skywork-OR1 dataset~\cite{skywork-or1-2025}, but with different difficulty filtering strategies. Specifically, for training the teacher model, we discard samples where the pass rate of \textit{R1-Distill-Qwen-1.5B} is lower than 25\%. For student training, we adopt a more tolerant threshold of 50\% to ensure adequate learning signals. We also filter based on prompt length, resulting in 58k and 46k samples for teacher and student training, respectively. As shown in Table~\ref{table:teacher} and Table~\ref{table:zero}, these difficulty-adjusted data support effective R1-Zero-like training.

\textbf{Implementation Details.}
Following prior efforts~\cite{zeng2025simplerlzooinvestigatingtamingzero}, we solely use the rule-based accuracy reward $r_{\text{acc}}$ for reinforcement learning to avoid negative effects from format constraints. 
The prompt is formatted as \texttt{"\{question\}\textbackslash nPlease reason step by step, and put your final answer within \textbackslash boxed\{\}."}, and is wrapped using the official chat template of the Qwen2.5 model family.
To accommodate the Zero-RL setting, we set the context length to 8K and the sampling temperature to 1.0. Moreover, for KD, we apply a constant KL coefficient of 5e-3, while for KDRL-Annealing, we set $\beta_{\text{init}}$ to 2e-1, $\delta$ to 8e-4, and $\beta_{\text{min}}$ to 2e-3. All Qwen2.5-3B models are trained for 480 steps. Additional hyper-parameters are summarized in Table~\ref{tab:zeroparams}.

\begin{table}[H]
\caption{Performance of the constructed teacher model in R1-Zero-like training. The teacher is obtained by applying GRPO to Qwen2.5-7B for 300 training steps.}
\label{table:teacher}
\centering
\footnotesize
\begin{tabular}{@{}p{3.3cm}@{}C{1.5cm}@{}C{1.5cm}@{}C{1.5cm}@{}C{1.5cm}@{}C{1.6cm}@{}C{1.6cm}@{}C{1.2cm}@{}}
\toprule
\multirow{2}{*}{\textbf{Model}} & \multirow{2}{*}{\textbf{AIME24}} & \multirow{2}{*}{\textbf{AIME25}} & \textbf{MATH} & \multirow{2}{*}{\textbf{AMC23}} & \textbf{Minerva} & \textbf{Olympiad} & \multirow{2}{*}{\textbf{Avg}}\\
 &  &  & \textbf{500} &  & \textbf{Math} & \textbf{Bench} &  \\
\midrule
Qwen2.5-7B              & 8.8  & 2.7 & 56.0 & 36.6 & 31.6 & 25.6 & 26.9 \\
Qwen2.5-7B-GRPO         & 17.1 & 11.5& 77.8 & 54.7 & 35.7 & 41.5 & 39.7 \\
\bottomrule
\end{tabular}
\vspace{-2mm}
\end{table}

\begin{table}[H]
\centering
\caption{Training hyper-parameters used in R1-Zero-like training. GRPO and KD-RKL baselines
share the same settings as KDRL, except for KL-related differences.}
\label{tab:zeroparams}
\begin{tabular}{lcc}
\toprule
\textbf{Hyper-parameter} & \textbf{Teacher Construction} & \textbf{KDRL Training} \\
\midrule
Train batch size & 64 & 64 \\
PPO mini-batch size & 64 & 64 \\
Number of rollouts & 16 & 16 \\
Learning rate & 1e-6 & 1e-6 \\
Sampling temperature & 1.0 & 1.0 \\
Max prompt length & 1,024 & 1,024 \\
Max response length & 8,192 & 8,192 \\
Training steps & 300 & 480 \\
Entropy coefficient & 1e-3 & 1e-3 \\
KL coefficient $\beta$ & - & 2e-1 $\rightarrow$ 2e-3 \\
KL loss type & - & \texttt{mse} \\
\bottomrule
\end{tabular}
\end{table}

\end{document}